\documentclass[10pt,twocolumn,letterpaper]{article}

\usepackage{cvpr}              

\usepackage[T1]{fontenc}
\usepackage{tabularx}
\usepackage{amsmath}
\usepackage{float}
\usepackage{cuted}
\usepackage{multirow}

\renewcommand{\paragraph}[1]{\vspace{.5em}\noindent\textbf{#1.}}

\usepackage{xspace}
\newcommand{\paper}{Di3PO - Diptych Diffusion DPO for Targeted Improvements in Image Generation}

\definecolor{cvprblue}{rgb}{0.21,0.49,0.74}
\usepackage[pagebackref,breaklinks,colorlinks,allcolors=cvprblue]{hyperref}

\def\paperID{21526} 
\def\confName{CVPR}
\def\confYear{2026}

\title{\paper}

\author{Sanjana Reddy\\
Google \\
\and
Ishaan Malhi \\
Google DeepMind \\
\and
Sally Ma \\
Google DeepMind \\
\and
Praneet Dutta\thanks{Work done while at Google DeepMind}\\
Google DeepMind \\
}

\begin{document}
\maketitle
\begin{abstract}
Existing methods for preference tuning of text-to-image (T2I) diffusion models often rely on computationally expensive generation steps to create positive and negative pairs of images. These approaches frequently yield training pairs that either lack meaningful differences, are expensive to sample and filter, or exhibit significant variance in irrelevant pixel regions, thereby degrading training efficiency. To address these limitations, we introduce "\paper", a novel method for constructing positive and negative pairs that isolates specific regions targeted for improvement during preference tuning, while keeping the surrounding context in the image stable.  We demonstrate the efficacy of our approach by applying it to the challenging task of text rendering in diffusion models, showcasing improvements over baseline methods of SFT and DPO.
\end{abstract}    
\begin{figure*}[t]
    \centering
    \includegraphics[width=1.0\linewidth]{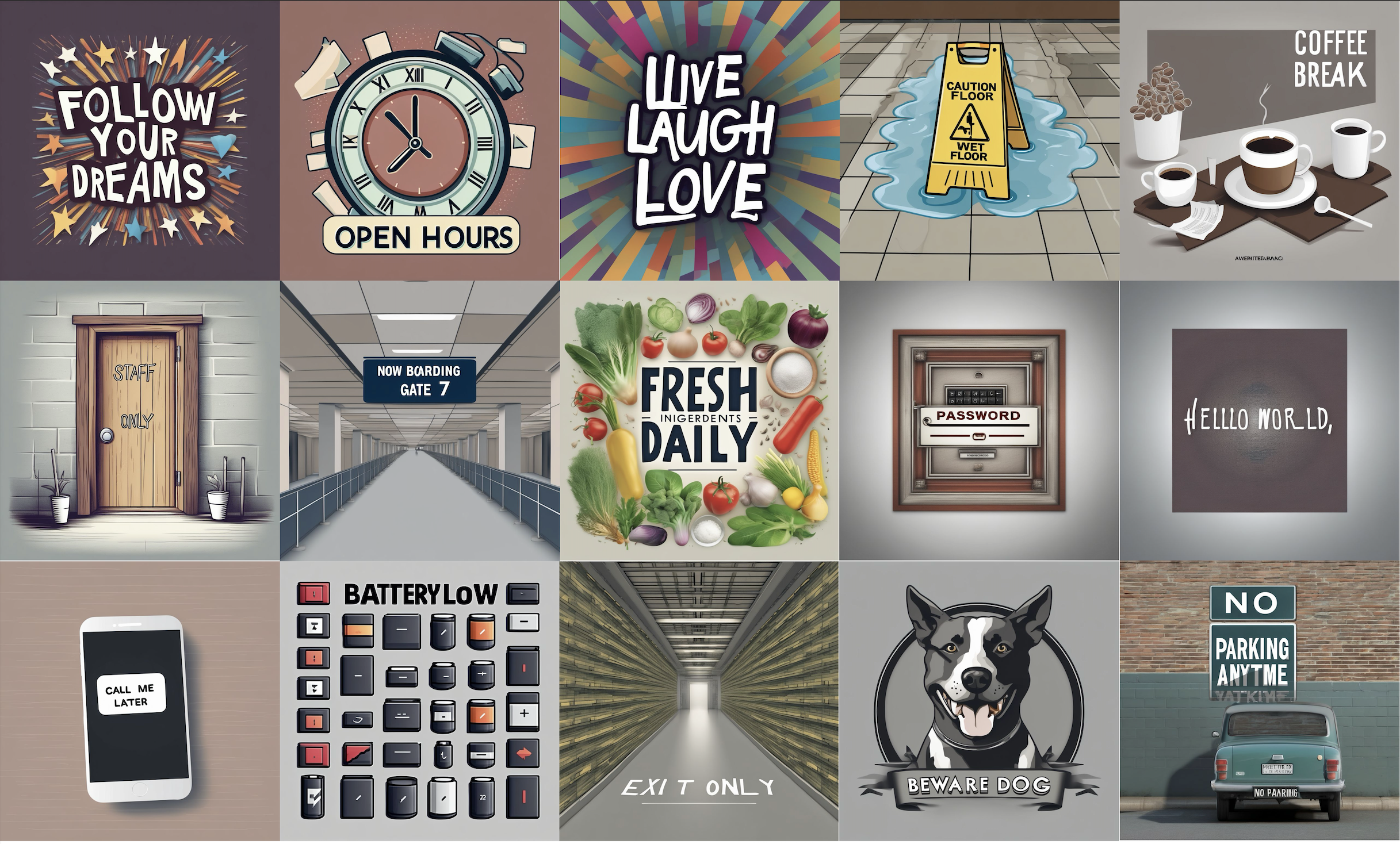}
    \caption{We develop \paper - a method to create DPO pairs for image preference tuning with minimal background changes. After finetuning on SDXL-1.0 on a Diptych set targeted for text rendering, our model demonstrates improved text rendering accuracy.}
    \label{fig:header_samples}
\end{figure*}

\section{Introduction}
\label{sec:intro}

Recent advancements in large-scale text-to-image (T2I) diffusion models  (\cite{sohl2015deep}\cite{song2020score}\cite{nichol2021improved}\cite{ho2020denoising}\cite{dhariwal2021diffusion}) showcase an ability to create high quality images from text prompts (\cite{ramesh2022hierarchical}\cite{rombach2022high}\cite{saharia2022photorealistic}), enabling the creation of highly photorealistic and creatively diverse visuals from natural language prompts. Despite these overarching successes, state-of-the-art models still showcase a quality gap in hard tasks such as text rendering, people generation, prompt adherence, structured generation and realism.

While several algorithms focus on preference tuning T2I models with the right priors \cite{wallace2024diffusion}\cite{lee2025calibrated}, reward models \cite{wu2023human} \cite{kirstain2023pickapicopendatasetuser} and training strategies \cite{yang2024usinghumanfeedbackfinetune}\cite{clark2024directlyfinetuningdiffusionmodels}, they often rely on creating large sets of preference pairs either by human grading or by rejection sampling using reward models. These approaches can often be computationally expensive as noted in \cite{yang2024usinghumanfeedbackfinetune}\cite{hiranaka2024hero} and seldom create strong preference pairs/sets that are sample efficient.

Existing preference tuning methods have been applied for tuning image generation models for human preferences \cite{esser2024scaling} to improve general visual quality. One such capability that is yet to show significant improvements is text rendering. State-of-the-art models persistently struggle with the precise rendering of legible and coherent text, frequently exhibiting failures such as "glyph splitting," misspelling, and inconsistent styling as observed in \cite{chen2023textdiffuserdiffusionmodelstext}\cite{chen2023diffuteuniversaltextediting}. This deficiency presents a critical bottleneck for practical applications in graphic design, where the seamless integration of text and visuals is essential. Addressing these failure modes is therefore important for advancing the utility of T2I models in professional workflows, necessitating more targeted and efficient optimization strategies than those currently available. The difficulty lies not only in accurately rendering individual characters and words but also in ensuring they are blended well with broader visual context. 

High-quality datasets play a pivotal role in training and evaluating T2I models on the task of text rendering. Existing methods have modified the architecture of diffusion models, either by adding more character tokens \cite{lu2025easytextcontrollablediffusiontransformer}, using additional parameterization \cite{Shen_2025_ICCV} or modifying the text encoder\cite{liu2024glyph} to process text correctly. At the same time, existing preference tuning methods often prefer to align diffusion models to human preferences\cite{kirstain2023pickapicopendatasetuser}\cite{yang2024usinghumanfeedbackfinetune}. We propose a simple training strategy to improve text rendering, building upon preference tuning methods where preference pairs are easily verifiable, either by OCR, or by construction. While we pick text rendering as a motivating example of applying our proposed method, it can be transferred to other tasks that are at the forefront of state-of-the-art image generation, which can be improved using preference tuning.

Diptych prompting \cite{shin2025largescaletexttoimagemodelinpainting}\cite{shin2025largescaletexttoimagemodelinpainting}\cite{hui2024hq} has emerged as a powerful capability in state-of-the-art image generation models. Existing prompting techniques leverage the capability of these models to create 2-panel images where minor changes can exist between the images within each panel. These have been leveraged successfully in the past to improve image editing capabilities. Diptych prompting relies on the in-context generation abilities of diffusion models\cite{huang2024incontextloradiffusiontransformers}\cite{shin2025largescaletexttoimagemodelinpainting}\cite{wu2025lesstomoregeneralizationunlockingcontrollability}. However, these techniques have yet to be explored for creating fine grained preference pairs. Building upon this line of research, in this work we showcase our novel method to create high quality preference pairs where one image in the panel is a positive pair, and the other image is carefully constructed to be a negative pair while keeping the context of the image minimally changed.

\section{Background and Related Work}

\subsection{Diffusion Models for Image Generation}
Denoising Diffusion Probabilistic Models (DDPMs) \cite{ho2020denoising} and their subsequent improvements \cite{nichol2021improved} have become the state-of-the-art framework for high-fidelity image generation. These models operate by progressively reversing a forward noising process. By training a model (typically a U-Net or DiT \cite{peebles2023scalablediffusionmodelstransformers}) to predict and remove Gaussian noise at each step, they can generate a coherent image from pure noise, conditioned on an input such as a text prompt. While exceptionally powerful, the outputs of these foundational models can still misalign with complex human intentions, necessitating further fine-tuning.

\subsection{Preference Tuning for Alignment}
To better align generative models with human aesthetics, safety, and instruction-following, preference tuning has been adapted from large language models. A leading method in this space is \textbf{Diffusion-DPO} \cite{wallace2024diffusion}, which applies Direct Preference Optimization to diffusion models. Diffusion-DPO, and related methods, align the diffusion models using human rated comparison data, training the model to increase the likelihood of a "preferred/winning" image while decreasing the likelihood of a "dispreferred/losing" image from a given pair.
A primary challenge for such methods, however, is the quality of the preference dataset. The effectiveness of DPO is reduced because of "visual inconsistency" \cite{hu2025d}, where significant visual disparities between the "good" and "bad" images in a pair - such as different backgrounds or compositions - prevent the model from identifying the specific factors that contribute to the preference. This introduces confounding signals and wastes expensive generation and labeling budgets. Other preference tuning methods \cite{lee2025calibrated} \cite{karthik2025scalable} have emerged to address the data and labeling cost, using multiple reward models to approximate human preferences and calibrate rewards without human annotation, but the challenge of generating minimally different pairs remains.

\subsection{High-Fidelity Editing and Diptychs}
The core problem of creating "minimal" preference pairs (e.g., an image with bad text vs.\ the \textit{exact same} image with good text) is analogous to the goal of high-fidelity, instruction-guided image editing. This field aims to make specific, localized changes to an image while preserving the background and all other content.

Prior works \cite{hui2024hq} \cite{xu2025insightedit} in this direction leverage large multimodal models to generate editing center datasets, and train downstream diffusion models to understand an edit instruction and apply it to a specific region, showing improvements in maintaining background consistency. These methods produce a "before" (original) and "after" (edited) image, which function as a conceptual Diptych. While these models are designed for user-facing editing, their ability to generate high-fidelity, minimal-change pairs provides a direct inspiration for creating the targeted positive/negative training data required for sample efficient preference tuning.

\section{Method}
\label{sec:method}

\subsection{Preliminaries}

To rigorously motivate Diptych Prompting, we examine the standard training objectives for diffusion models and their adaptation for preference optimization. We show that minimizing visual differences in training pairs directly optimizes the gradient signal in the DPO objective.

\subsubsection{DDPM and Diffusion DPO}
Standard Denoising Diffusion Probabilistic Models (DDPM) are trained using a simplified mean squared error objective. For an image $x_0$, timestep $t$, and noise $\epsilon \sim \mathcal{N}(0, \mathbf{I})$, the loss is:
\begin{equation}
\mathcal{L}_{\text{DDPM}}(\theta) = \mathbb{E}_{x_0, t, \epsilon} \left[ \| \epsilon - \epsilon_\theta(x_t, t, c) \|^2 \right]
\end{equation}
where $x_t = \sqrt{\bar{\alpha}_t}x_0 + \sqrt{1-\bar{\alpha}_t}\epsilon$ is the noisy latents, $c$ is the conditioning (e.g., text prompt), and $\epsilon_\theta$ is the denoising network.

The loss for Diffusion-DPO is formulated as

\begin{equation}
\mathcal{L}_{\text{DPO}}(\theta) = - \mathbb{E}_{(x_w, x_l), t, \epsilon} \left[ \log \sigma \left( \beta \left( \Delta(x_w) - \Delta(x_l) \right) \right) \right]
\label{eq:dpo}
\end{equation}
where $\Delta(x)$ represents the relative improvement of the active policy over the reference model for a given image:
\begin{equation}
\Delta(x) = \| \epsilon - \epsilon_{\text{ref}}(x_t, t, c) \|^2 - \| \epsilon - \epsilon_\theta(x_t, t, c) \|^2
\end{equation}
Crucially, standard implementations sample the \textit{same} timestep $t$ and noise $\epsilon$ for both $x_w$ and $x_l$ in a pair to reduce variance.

The core hypothesis of our approach is that increasing the visual similarity between positive ($x_w$) and negative ($x_l$) training pairs—outside of the specifically targeted improvement region—significantly enhances the efficiency of preference learning. We attribute this improvement to two primary factors: the mitigation of the credit assignment problem and the maximization of relevant gradient signal.

\subsubsection{Mitigating the Credit Assignment Problem}
In standard preference tuning setups, preference pairs are often generated via distinct random seeds or vastly different base models to ensure a noticeable quality gap. While this successfully creates a preferred and dispreferred image, it introduces severe confounding variables. If $x_w$ differs from $x_l$ in background composition, lighting, \textbf{and} text rendering, the optimization process would struggle to assign "credit" for the preference label correctly. The model may inadvertently learn to prefer specific global compositions or local features rather than improved text fidelity, or worse, learn unrelated patterns present in winning images.

Diptych Prompting explicitly resolves this by fixing the majority of the image context. By ensuring that $x_w \approx x_l$ for all distinct regions $R$ not containing the target features (e.g., text glyphs), we remove these confounding signals. The only discriminative features available to the objective function are those we intend to optimize.

\subsubsection{Precise Gradient Targeting \& Signal Concentration}
The benefit of Di3PO becomes evident when analyzing the gradient of the DPO loss with respect to the model parameters $\theta$. Let $w(x_w, x_l)$ be the scalar weight derived from the sigmoid term. The gradient is approximately:
\begin{multline}
\nabla_\theta \mathcal{L}_{\text{DPO}} \approx - w(x_w, x_l) \nabla_\theta \| \epsilon - \epsilon_\theta(x_{w,t}, t, c) \|^2 \\  - \nabla_\theta \| \epsilon - \epsilon_\theta(x_{l,t}, t, c) \|^2 
\label{eq:gradient}
\end{multline}
In a Diptych pair, $x_w$ and $x_l$ are identical for all pixels in the background region $R_{bg}$. Since standard Diffusion DPO uses identical noise $\epsilon$ for the pair, the noisy latents are also identical in this region: $x_{w,t}[i,j] = x_{l,t}[i,j]$ for $(i,j) \in R_{bg}$.

Consequently, for the spatial features corresponding to $R_{bg}$, the denoising network $\epsilon_\theta$ produces nearly identical outputs and gradients. In Equation \cref{eq:gradient}, these identical terms cancel out:
\begin{equation}
\nabla_\theta \text{Loss}_{R_{bg}}(x_w) \approx \nabla_\theta \text{Loss}_{R_{bg}}(x_l) \implies \nabla_{\theta_{R_{bg}}} \mathcal{L}_{\text{DPO}} \approx 0
\end{equation}

This cancellation naturally focuses the entire magnitude of the gradient update on the parameters responsible for the differing regions (the text). By explicitly constructing pairs where $R_{bg}$ is maximized, Di3PO ensures that model capacity is not wasted attempting to differentiate between irrelevant background noise, thereby increasing the signal-to-noise ratio of the preference learning update.

The DPO objective increases the log-likelihood of preferred data while decreasing it for dispreferred data. When $x_w$ and $x_l$ share identical pixel values across most of the image, the gradient contributions from these shared regions largely cancel out during the KL derived update step. Consequently, the magnitude of the gradient updates becomes highly concentrated in the regions where the images differ.

In our application to text rendering, this means the model's parameters are updated almost exclusively based on the difference between malformed glyphs in $x_l$ and correct glyphs in $x_w$, rather than wasting model capacity adjusting unrelated background textures. This results in a higher signal-to-noise ratio during training, allowing for faster convergence with fewer distinct training pairs. Using the theoretical motivation as the basis of our experiments, we now show the method to create the datasets, along with experimental results.

\subsection{Diptych Preference Pair Generation}

Our methodology for generating preference pairs is designed to be both scalable and targeted, creating high-fidelity diptychs that isolate the specific attribute we aim to improve—in this case, text rendering. Our method has key advantages compared to existing approaches:

\begin{enumerate}
    \item Diptych Prompting relies on sampling the same model with in-context generation capabilities. This ensures the background of the image is very close for both positive and negative pairs.
    \item We do not require any task specific rewards (Human Preference Scores, Preference Reward models etc), and instead create preference pairs via construction, making the process simpler, reward model free and computationally cheap.
    \item All preference pairs can be generated offline and do not require expensive online sampling during RL training.
\end{enumerate}

However, our method also has a few limitations:

\begin{enumerate}
    \item Our method relies on the base T2I model having the ability to create Diptych pairs. We observed that SDXL 1.0 in itself cannot generate Diptych pairs, so we opt to use state-of-the-art T2I models that exhibit this property as shown in \cref{fig:diptych_image}, following the observation in \cite{hui2024hq}. While this makes the target images of a possibly higher fidelity/quality, we demonstrate how our approach still performs better than regular SFT on winning images.
    \item Our method assumes mispellings to be "in-distribution" for the base T2I model, i.e the T2I model is more likely to generate mispellings closer to our chosen negative image, than the correct spelling in the positive pair. Our negative image prompts are chosen based on empirical evidence observed by SDXL 1.0, where it is prone to repeat or omit characters in the text seeking prompt. This loss pattern can be specific to the model being trained, and constructing the right losing image is an important future direction.
\end{enumerate}

Our process can be broken down into two main stages: data generation and data filtering.

\subsubsection{Data Generation}

\begin{figure}[H]
    \centering
    \includegraphics[width=0.9\linewidth]{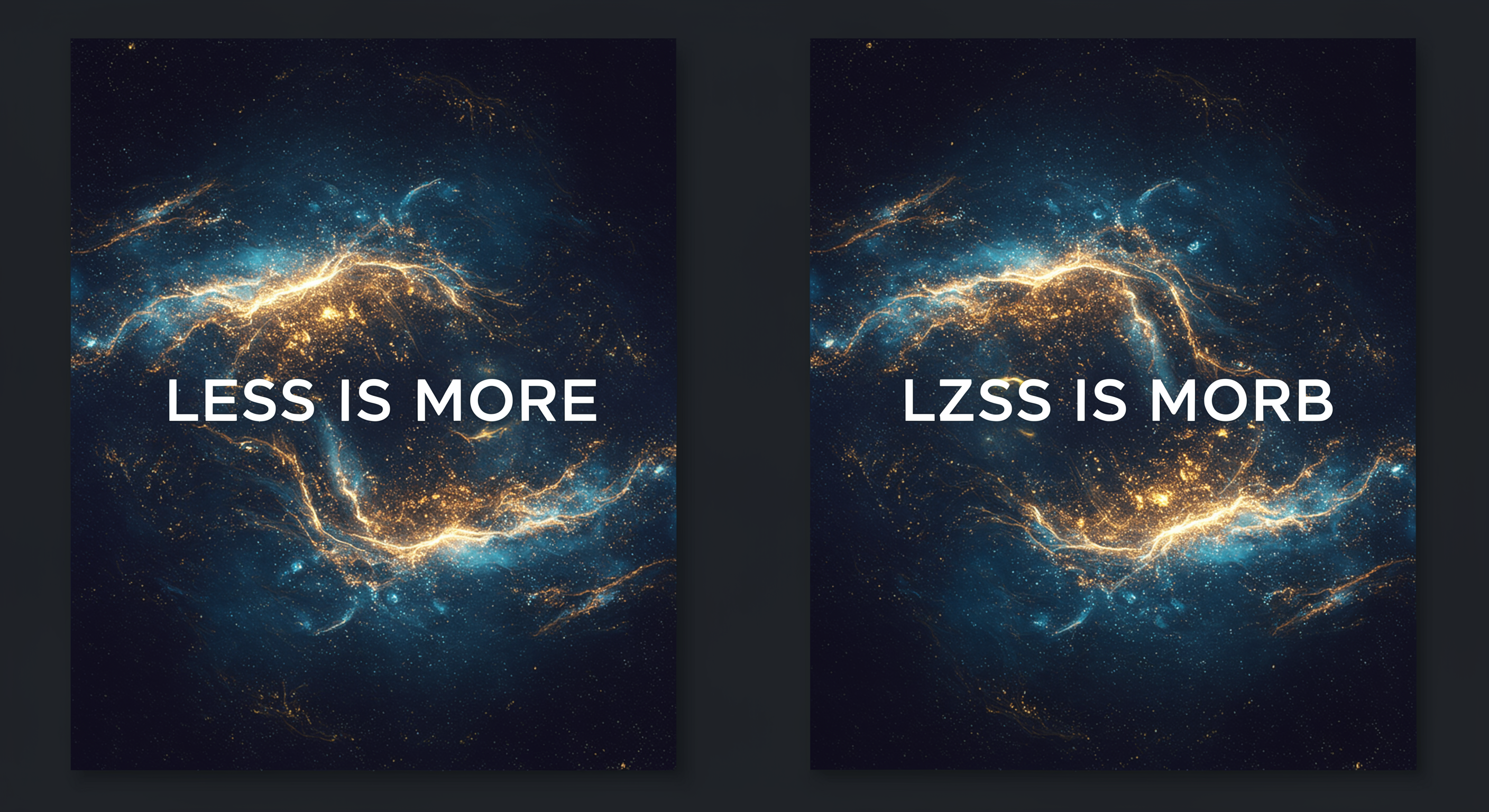}
    \caption{Dipytch Image generated using one single prompt, generating Dipytch images ensures consistency of the background, allowing the model to focus on the text rendering}
    \label{fig:diptych_image}
\end{figure}

\begin{figure}[H]
    \centering
    \includegraphics[width=0.45\linewidth]{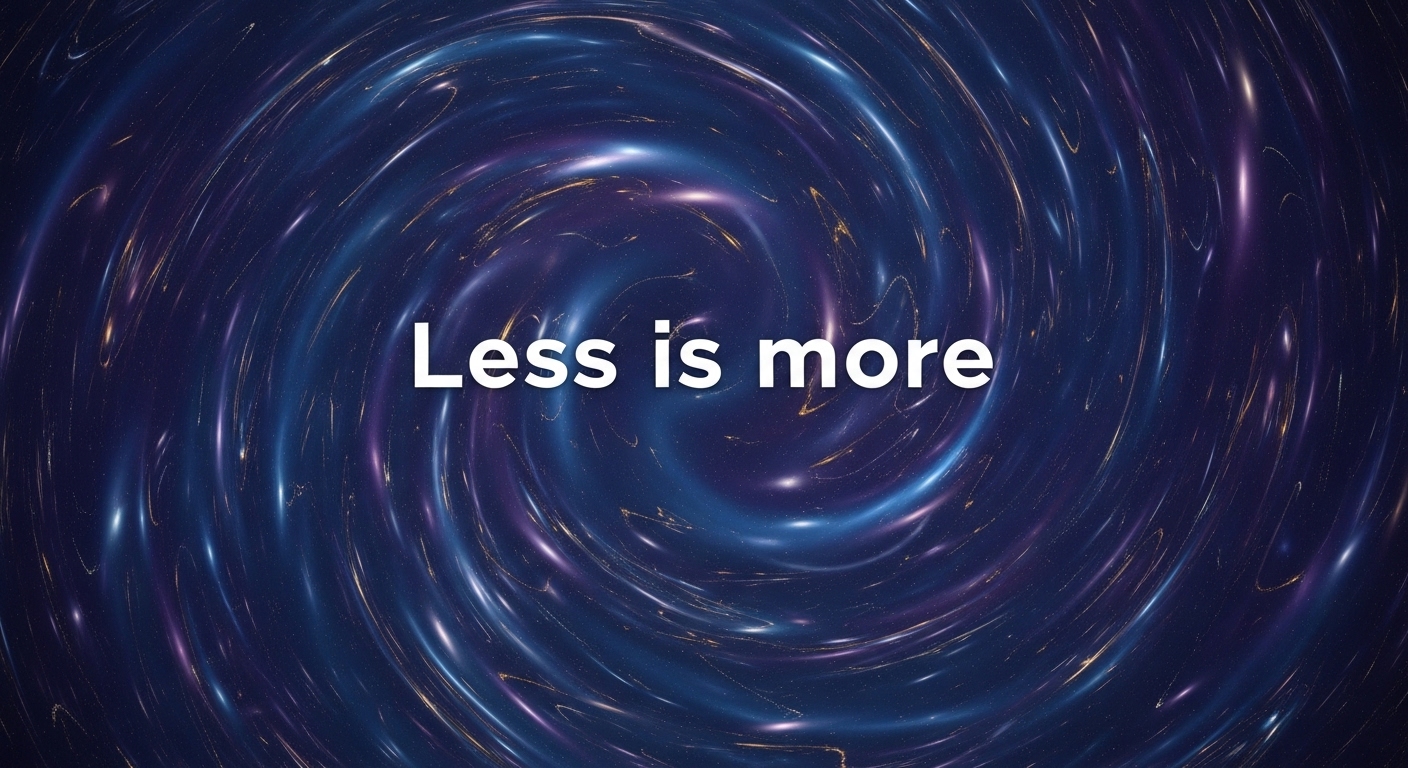}
    \includegraphics[width=0.45\linewidth]{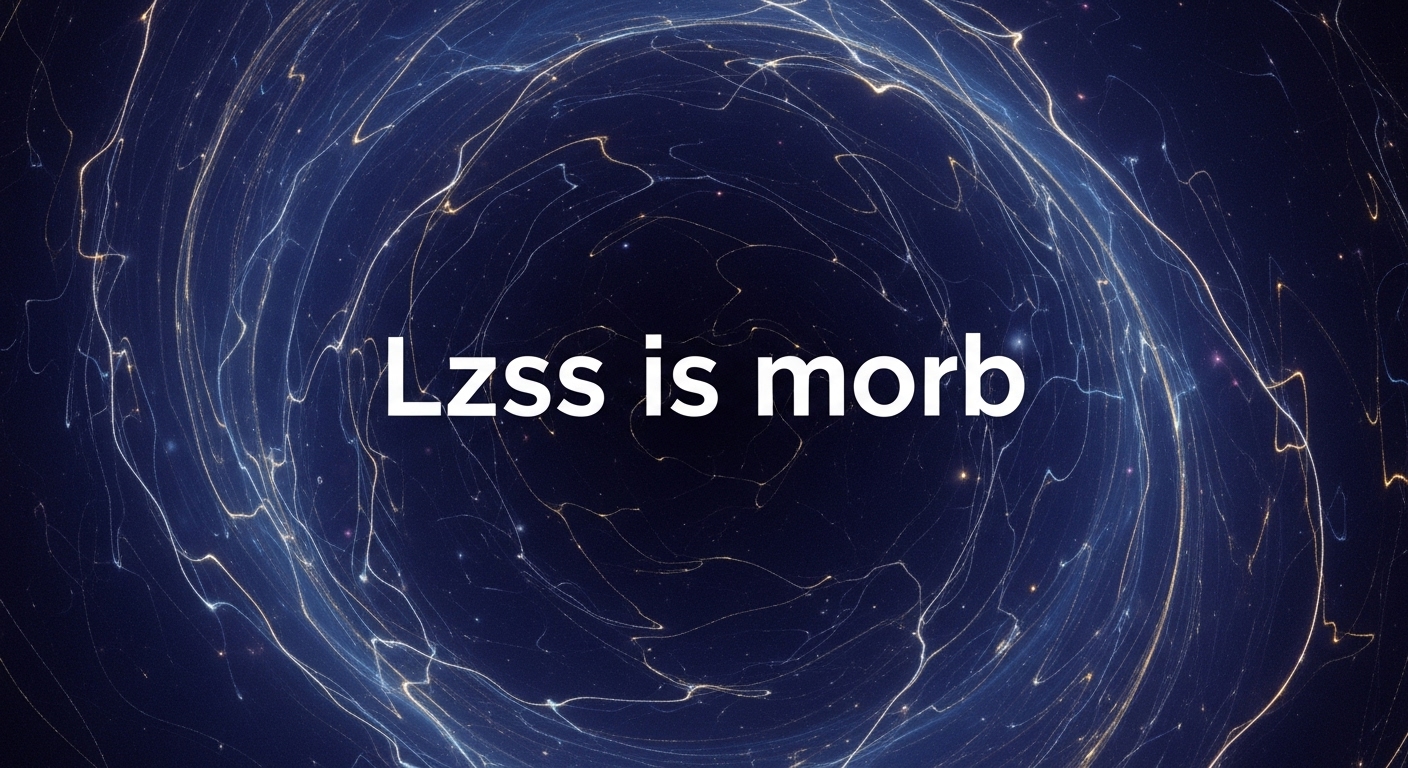}
    \caption{Winning and losing image generated using two separate prompts results in different backgrounds}
    \label{fig:winning_losing}
\end{figure}

The data generation process as shown in \cref{fig:data_pipeline_workflow} is a multi-step pipeline that leverages large language and image generation models to create a dataset of preference pairs as shown in \ref{fig:diptych_image}. The workflow is as follows:

\begin{figure*}[!th]
  \centering
  \includegraphics[width=0.8\textwidth]{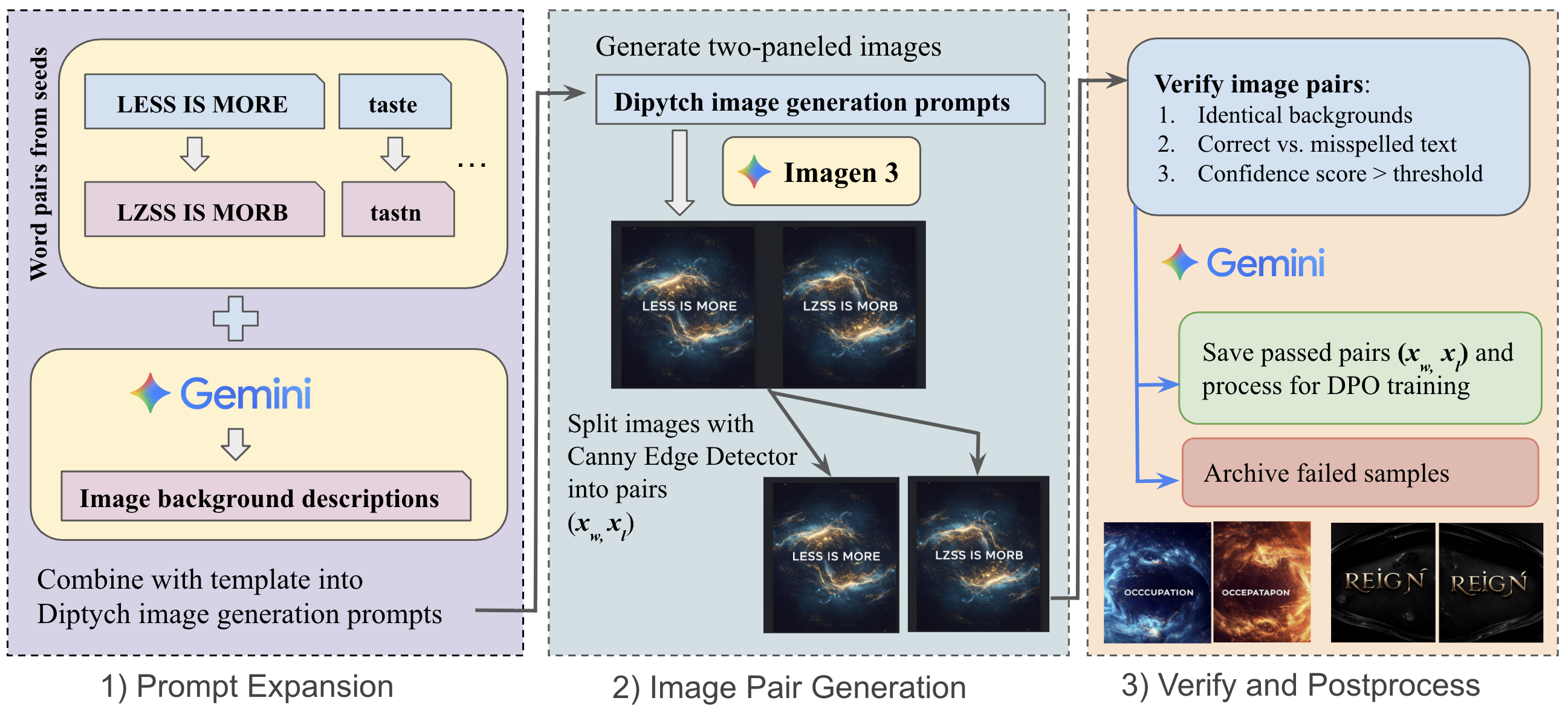}
  \caption{Preference pair generation workflow for creating Diptych pairs for DPO tuning.}
  \label{fig:data_pipeline_workflow}
\end{figure*}

\begin{enumerate}
    \item \textbf{Input Seed Data:} We begin with a seed set of correctly spelled words. For each word, we programmatically generate a corresponding misspelled version by randomly modifying 20\% of its characters. This approach of providing explicit misspellings was found to be more effective than prompting a large language model to generate them. This process results in pairs of \textit{\{correct\_word, misspelled\_word\}}, which form the basis for our preference pairs.

    \item \textbf{Background Generation:} For each word pair, we use a large language model (Gemini 2.5) \cite{comanici2025gemini25pushingfrontier} to generate a detailed and creative description of a background scene. The model is prompted to act as a graphic designer, ensuring the generated backgrounds are diverse and high-quality.

    \item \textbf{Diptych Prompting:} The generated background description is then combined with a prompt template that instructs the image generation model to create a two-panel (Diptych) image \cref{fig:diptych_image}. The prompt explicitly requests the same background in both panels, with the correctly spelled word rendered in one panel and the misspelled word in the other. This ensures that the only significant difference between the two images is the quality of the rendered text. See \cref{sec:appendix_prompts} for more details on this prompt.

    \item \textbf{Image Generation and Splitting:} The composed prompt is used to generate a single, wide landscape image containing the two panels side-by-side. This diptych is then split into two separate images. The split is performed using Canny edge detection \cite{canny} to identify the border between the panels, with a fallback to a heuristic-based middle split if the edge detection is not confident. The left panel, containing the correctly spelled word, is designated as the "winning" image ($x_w$), and the right panel, with the misspelled word, as the "losing" image ($x_l$).
\end{enumerate}

\subsubsection{Data Filtering}

To ensure the quality of our training data, we employ a rigorous filtering process that validates each generated image pair:

\begin{enumerate}
    \item \textbf{Automated Verification:} We use a multimodal model (Gemini 2.5) to verify each pair of generated images. The model is prompted to act as a human rater and check for two main criteria: 1) the backgrounds of the two images are identical, and 2) the text in the two images is slightly different, with both images containing text.

    \item \textbf{Confidence Scoring:} The verification model provides a `passing' decision (True/False) and a `confidence' score (0-100) for each pair. We consider a pair as "passed" only if the `passing' decision is True and the confidence score is above a certain threshold (e.g., >70). This ensures a high level of precision in our final dataset. See \cref{sec:appendix_prompts} for additional details around verification prompts.
\end{enumerate}

This two-stage process allows us to generate a large-scale, high-quality dataset of preference pairs for fine-tuning text-to-image models. To demonstrate the sample efficiency of our approach, we generate 300 image pairs with text seeking prompts for our experiments.

\section{Experiments}
\subsection{Models}
We use Stable Diffusion XL (SDXL)\cite{podell2023sdxlimprovinglatentdiffusion} and SD3 as our base T2I models. The training set was created by collecting 300 Diptych pair samples from a synthetic text rendering prompt dataset for the text rendering task by sampling from the Imagen 3 API \cite{imagenteamgoogle2024imagen3} \footnote{\url{https://fal.ai/models/fal-ai/imagen3} or \url{https://ai.google.dev/gemini-api/docs/imagen}}.  This prompt dataset was collected by prompting Gemini 2.5 \cite{comanici2025gemini25pushingfrontier} to give us text rendering prompts for images.  Additional details on prompts for dataset creation are available in \cref{sec:appendix_prompts}, along with some examples of Diptych pairs in \cref{fig:passed_example}.

In our experimental evaluation, we assess the performance of the SDXL 1.0 model \cite{wallace2024diffusion} when fine-tuned using our proposed Di3PO method. To provide a comprehensive comparison, we establish two baselines: the original, pre-trained SDXL 1.0 and SD3 models available on Hugging Face\footnote{\url{https://huggingface.co/docs/diffusers/en/using-diffusers/sdxl}}, and a supervised fine-tuned (SFT) version of SDXL 1.0 that was trained exclusively on the "winning" images from our Diptych preference pairs.

\begin{figure}[H]
    \centering
    \includegraphics[width=1.0\columnwidth]{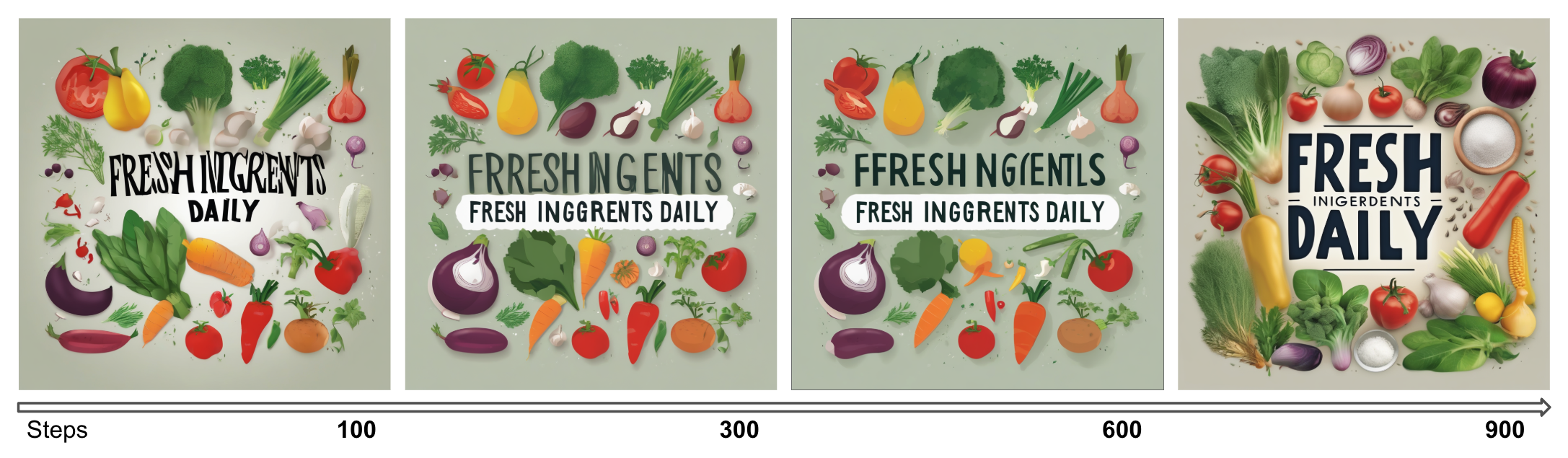}
    \caption{Image generation from various checkpoints during the tuning process}
    \label{fig:checkpoints}
\end{figure}

For training, we utilized a TPU v4 slice equipped with 8 chips. The DPO fine-tuning was conducted for 900 steps with a learning rate of $3 \times 10^{-8}$ with the Adam optimizer \cite{kingma2017adammethodstochasticoptimization} and a batch size of 16. All image samples used for metric calculation were generated by running the model for 50 diffusion steps with a guidance value of 7.5. The progressive improvement in text rendering quality throughout the training process can be observed in the checkpoint samples presented in \cref{fig:checkpoints}.

\subsection{Datasets}

We use the Diptych generation meta-prompt that leverages a state-of-the-art LLM to generate a comprehensive suite of text-seeking prompts. For each prompt intent, we first manually crafted a small collection of "golden prompts" that clearly exemplify the targeted text rendering capability. These golden prompts were iteratively refined through internal testing. We then used these golden prompts as examples within a meta-prompt framework, instructing a state-of-the-art LLM to generate hundreds of additional prompts that maintain the same core intent but introduce varied scenarios, difficulty levels, and linguistic structures. Each generated prompt was then validated through a multi-stage process to ensure quality, removing any prompts that inadvertently tested multiple intents or failed to properly isolate the target capability.

For the training dataset, we use the text in the prompt and it's misspelling as seeds in our Diptych Pipeline to generate 300 prompts for training.
For the evaluation dataset, we use base prompts generated by the LLM as inputs for evaluating text rendering. Data statistics are shown in \cref{tab:eval_stats}. Additional details on the meta-prompt for the prompt generation, verification and passing/failing samples are detailed in \cref{sec:appendix_prompts}, along with samples of evaluation data in \cref{sec:appendix_eval_data}.

\begin{table}[H]
    \centering
    \begin{tabular}{cc}
    \hline
    Average Character Count& 65.61 \\
    Average Word Count & 10.65 \\
    Average Ground Truth Character Count& 26.66 \\
    \hline
    \end{tabular}
    \caption{Evaluation Data Statistics}
    \label{tab:eval_stats}
\end{table}

\section{\textbf{Results}}

\begin{figure*}[ht]
    \centering
    \includegraphics[width=0.95\linewidth]{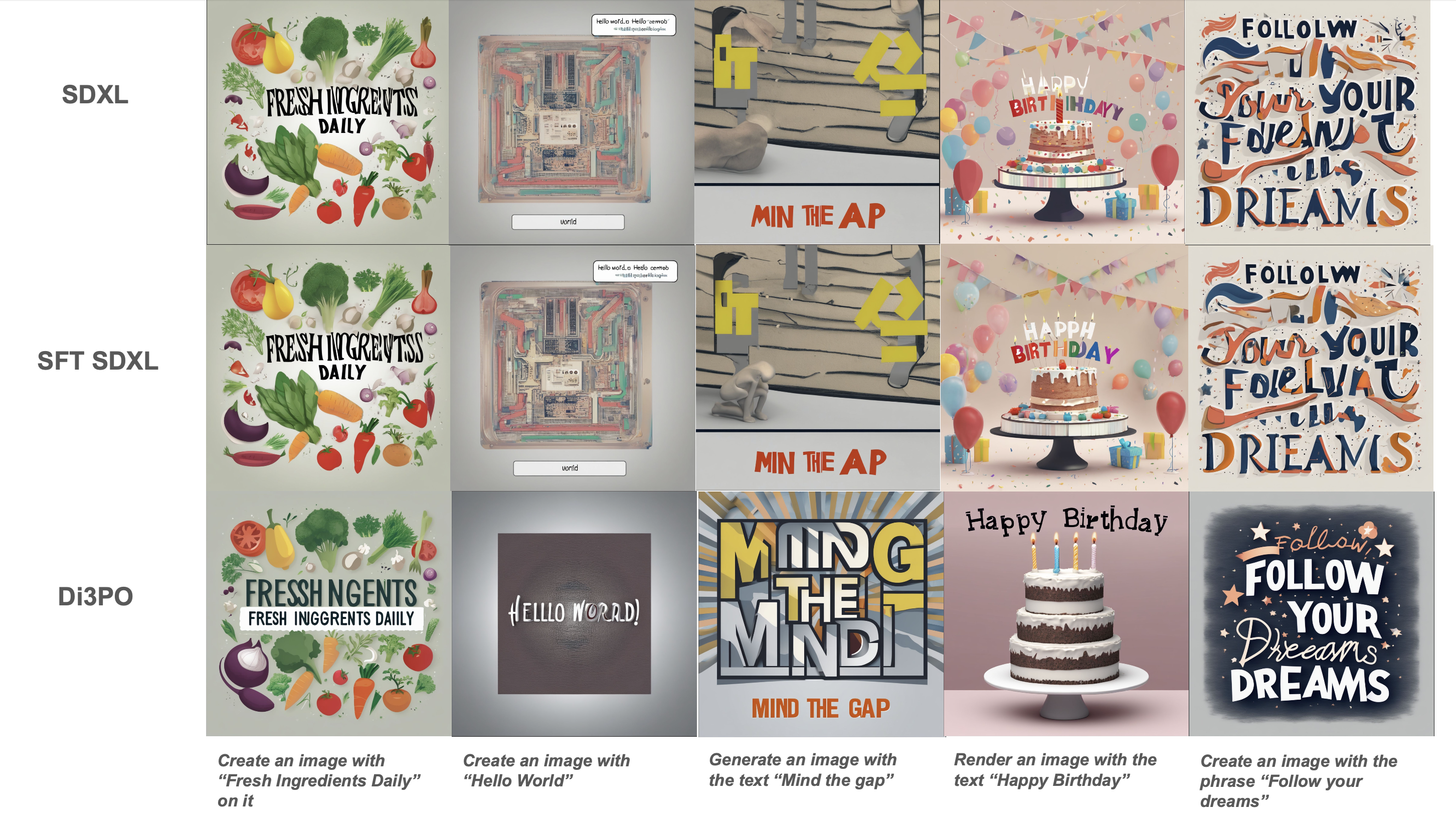}
    \caption{Comparison of images generated for prompts by pre-trained SDXL 1.0, SFT tuned SDXL and Di3PO tuned SDXL}
    \label{fig:comparison}
\end{figure*}

Our empirical results indicate that the Di3PO method leads to a significant reduction in text rendering error rates when compared to the SFT baseline, as detailed in \cref{tab:gocr_results}. A key advantage of our approach is its efficiency; it enhances text rendering without the need for auxiliary reward functions, computationally expensive diffusion sampling, or a large number of fine-tuning samples.

We also observe that SFT finetuning on the winning image, which in this case is similar to distilling from an external model, is prone to model collapse where the model's learning curve becomes noisy after a few 100 steps.  We provided qualitative samples in \cref{fig:comparison} to illustrate the improved text rendering performance compared to the baselines.

\subsection{{\textbf{Evaluations}}}

We leverage existing OCR (Optical Character Recognition) APIs \footnote{\url{https://docs.cloud.google.com/vision/docs/ocr}} to measure the accuracy of the text generated in the images. We compare the OCR-detected text against the ground truth from our 2000 evaluation prompts, which were not used during the DPO/SFT tuning phase. Our evaluation relies on the following metrics: Levenshtein edit distance, Word Error Rate, and Substring Match Ratio. We sample images with 4 different seeds and report average values of the metrics on 1 seed, and average and best metrics over 4 seeds.

\begin{table}[htbp]
\centering
\resizebox{\columnwidth}{!}{%
\begin{tabular}{l c c c} 
\toprule
\textbf{Model} & \textbf{Edit Distance $\uparrow$} & \textbf{Word Error Rate $\downarrow$} & \textbf{Substring Match Ratio $\uparrow$} \\
\midrule

Pretrained SDXL 1.0 & & & \\ 
\quad Average (n = 4) & $0.1305 \pm \scriptstyle 0.0054$ & $0.7215 \pm \scriptstyle 0.0117$ & $0.0619 \pm \scriptstyle 0.0084$ \\
\quad BoN (n = 4) & $0.2678 \pm \scriptstyle 0.0151$ & $0.4677 \pm \scriptstyle 0.0253$ & $0.1736 \pm \scriptstyle 0.0263$ \\
\addlinespace
\midrule

SFT SDXL 1.0 & & & \\
\quad Average (n = 4) & $0.1290 \pm \scriptstyle 0.0056$ & $0.7213 \pm \scriptstyle 0.0110$ & $0.0613 \pm \scriptstyle 0.0079$ \\
\quad BoN (n = 4) & $0.2640 \pm \scriptstyle 0.0146$ & $0.4744 \pm \scriptstyle 0.0243$ & $0.1715 \pm \scriptstyle 0.0255$ \\
\addlinespace
\midrule

Baseline DPO SDXL 1.0 (Background variation) & & & \\ 
\quad Average (n = 4) & $0.1288 \pm \scriptstyle 0.0050$ & $0.7446 \pm \scriptstyle 0.0110$  & $0.0455 \pm \scriptstyle 0.0094$ \\
\quad BoN (n = 4) & $0.2510 \pm \scriptstyle 0.0147$ & $0.5306 \pm \scriptstyle 0.0245$ & $0.1265 \pm \scriptstyle 0.02455$ \\
\addlinespace
\midrule

\textbf{Di3PO SDXL 1.0 (Ours)} & & & \\
\quad Average (n = 4) & \textbf{$0.1640 \pm \scriptstyle 0.0063$} & \textbf{$0.6456 \pm \scriptstyle 0.0124$} & \textbf{$0.0946 \pm \scriptstyle 0.0101$} \\
\quad \textbf{BoN (n = 4)} & \textbf{\boldmath$0.3353 \pm \scriptstyle 0.0160$} & \textbf{\boldmath$0.3826 \pm \scriptstyle 0.0231$} & \textbf{\boldmath$0.2506 \pm \scriptstyle 0.0285$} \\
\bottomrule

Pretrained SD3 & & & \\
\quad Average (n = 4) & $0.3794 \pm \scriptstyle 0.0064$ & $0.6764 \pm \scriptstyle 0.0110$ & $0.0850 \pm \scriptstyle 0.0080$ \\
\quad BoN (n = 4) & $0.6363 \pm \scriptstyle 0.0156$ & $0.4244 \pm \scriptstyle 0.0246$ & $0.2480 \pm \scriptstyle 0.0234$ \\
\addlinespace
\midrule

\textbf{Di3PO SD3 (Ours)} & & & \\
\quad Average (n = 4) & \textbf{$0.4734 \pm \scriptstyle 0.0170$} & \textbf{$0.5758 \pm \scriptstyle 0.0221$} & \textbf{$0.0955 \pm \scriptstyle 0.0255$} \\
\quad \textbf{BoN (n = 4)} & \textbf{\boldmath$0.7228 \pm \scriptstyle 0.0143$} & \textbf{\boldmath$0.3250 \pm \scriptstyle 0.0210$} & \textbf{\boldmath$0.3436 \pm \scriptstyle 0.0310$} \\
\bottomrule
\end{tabular}
}
\caption{Results for various text rendering metrics on trained models and baselines. Our method (Di3PO) shows lower word error rate, and higher substring match and edit distance values. We report metrics over N=4 random samples with average and Best-of-N (BoN) statistics}
\label{tab:gocr_results}
\end{table}

The Levenshtein edit distance is normalized and subtracted from 1.0, where a value closer to 1.0 indicates a better match between the generated text and the ground truth. For Word Error Rate, which calculates the number of word errors (substitutions, deletions, and insertions) divided by the total number of words, a lower value signifies better performance. The Substring Match Ratio measures the proportion of correctly rendered substrings, where a higher value is better.

To establish confidence intervals for these metrics, we employ a non-parametric bootstrap method with 1000 bootstrap replicas.

\subsection{{\textbf{Discussion}}}

Our results demonstrate that Di3PO significantly improves text rendering in the SDXL 1.0 model. As shown in \cref{tab:gocr_results}, our method outperforms both the base model and the SFT baseline across all three of our evaluation metrics. Specifically, we observed an increase in the Levenshtein Distance and Substring Match Ratio, alongside a decrease in the Word Error Rate. It is also worth noting that while these metrics show a clear quantitative improvement, there are also qualitative improvements in the generated images, as shown in \cref{fig:comparison}, which are not fully captured by the metrics alone.

One of the key advantages of our approach is its sample efficiency. The targeted nature of Diptych preference pairs allows for rapid learning without the need for a large number of fine-tuning samples. This is in stark contrast to the SFT baseline, which we observed to be prone to model collapse after only a few hundred training steps when using a small dataset. This suggests that for tasks requiring high precision with limited data, our method offers a more stable and effective training paradigm.

Our choice of metrics—Levenshtein edit distance, Word Error Rate, and Substring Match Ratio—was deliberate. These metrics provide a comprehensive evaluation of text rendering quality. Levenshtein distance measures the character-level accuracy, Word Error Rate captures mistakes at the word level, and the Substring Match Ratio assesses the model's ability to generate correct sequences of characters within the text. Together, they offer a nuanced view of the model's performance beyond simple pixel-level comparisons.

Looking ahead, there are several promising directions for future research. While this work focused on DPO, exploring other reinforcement learning algorithms for preference tuning could yield further improvements. Additionally, the Diptych preference pair generation method is not limited to text rendering. Applying this technique to other challenging, localized tasks in image generation—such as improving people generation, prompt adherence, structured generation and realism is an important future research direction.

\section{Conclusion}

In this work, we introduce \paper, a method for generating high-quality preference pairs for diffusion model alignment. By isolating specific regions of interest—such as text rendering—while maintaining pixel-perfect consistency in the surrounding context, our approach effectively mitigates the credit assignment problem inherent in standard Diffusion DPO. Our theoretical analysis and empirical results demonstrate that this precise gradient targeting significantly improves training efficiency and final model performance. Di3PO offers a scalable pathway for fine-grained control over generative models, moving beyond broad aesthetic tuning to address specific, localized failure modes in professional workflows.

\section*{Acknowledgment}
We sincerely thank Yinxiao Li for his technical guidance and for sharing the codebase that helped speed up our experimentation.
{
    \small
    \bibliographystyle{ieeenat_fullname}
    \bibliography{ref}
}

\clearpage
\setcounter{page}{1}
\appendix
\maketitlesupplementary
\begin{strip}
\section{Prompt Templates for Data Generation and Filtering}
\label{sec:appendix_prompts}

\subsection{Diptych Generation Prompt}
To generate the diptych images, we first use Gemini with a metaprompt instructing it to act as a graphic designer and generate a detailed background.
\begin{verbatim}
"""You are a graphic designer responsible 
for high-quality text based graphics. You need to write a prompt for 
a text to image model to generate backgrounds for text to be generated on. 
The prompt needs to be as detailed as possible. Describe the background 
on which you expect the text to appear. Include specific details of the 
background, such as its color, shape, material, texture, or style. 
Make sure your description is very clear and creative. Do not include 
{right_input} or {misspelling} in the background description.

Then output an answer in the following format:

generated_background: the background for the image to render, make sure 
not to include the word {right_input} or the word {misspelling} in the 
background description.

Here is an example of a good answer:

generated_background: The font of the text is an elegant font, such as a 
delicate script or a bold serif. The text should be the focal point of the 
image, with a size and placement that commands attention. For the background, 
create a scene of a serene, misty forest at dawn. The color palette 
should be soft and ethereal, with pale blues, greens, and pinks blending 
seamlessly. Sunlight filters through the mist, creating beams of light that 
illuminate the scene. Dewdrops cling to leaves and spiderwebs, catching the 
light and adding a touch of sparkle. The overall atmosphere should evoke a 
sense of tranquility and wonder. The word should be rendered in a color that 
complements the background, perhaps a warm gold or a shimmering silver. 
It should appear as if it's part of the scene, perhaps nestled among the 
leaves or rising with the mist. The text should have a subtle texture, 
perhaps resembling dewdrops or etched metal, to further integrate it with the 
background. This image should capture the essence of beauty in both its 
visual and textual form, creating a captivating and memorable graphic."""
\end{verbatim}

The output from Gemini (\texttt{generated\_background}) is then inserted into the following T2I prompt template to create the final prompt sent to the image generation model.

\begin{verbatim}
"""Two images with a left and right panel, placed
side by side. Both images are fundamentally identical in terms of their
backdrop, lighting, and color palettes.
The left and right panel have this background.

{first_orientation} Image: Create an image with this background below.
On this image render the word {right_input}. The word {right_input} is
placed on the same background as the second image.

{second_orientation} Image: Create an identical image to the first image
with the exact same background. The word {misspelling} is placed on the
same background as the first image. It is extremely important to spell
the word as **{misspelling}**.

Background: {generated_background}
"""
\end{verbatim}

\subsection{Verification Prompt}
For the filtering stage, we use Gemini with the following prompt to verify the quality and correctness of the generated pairs.

\begin{verbatim}
"""You are a rigorous human rater for text on 
image rendering graphics company. You are given two images, and you need 
to verify that they are identical. The first image should be showcasing 
text in the same background as the second image. You must check to make 
sure that the background is identical, and that the text is rendered in 
the same background as the second image. You must carefully attend to 
even tiny details to make sure every single detail of the background, 
such as color, shape, design, and style, is the same. You must also 
check to make sure that the text in the first image is only slightly 
different from the text in the second image. Both images should have 
text in them. But, the text should not be the same in both images.

Then output an answer in the following format:

explanation: thought process and statement to justify your decision
passing: true or false indicating whether both checks are passed or not
confidence: a confidence score in your above decision of passing, out of 100

Some examples of the output given two images are as follows:
explanation: "The text on both the images is the different but there are
minor differences in the background. The background has slightly
different color."
passing: true
confidence: 80

explanation: "The text in the images are the same but the background is
different."
passing: false
confidence: 10

explanation: "The text in the second image is completely missing."
passing: false
confidence: 0

explanation: "The text in both the images is different. The backgrounds
are the same and the text in both images has been rendered clearly."
passing: true
confidence: 100
"""
\end{verbatim}\end{strip}

\section{Qualitative Samples of Preference Pairs}
\label{sec:appendix_samples}
\begin{figure*}[h]
  \centering
  \begin{minipage}{0.48\textwidth}
    \includegraphics[width=\linewidth]{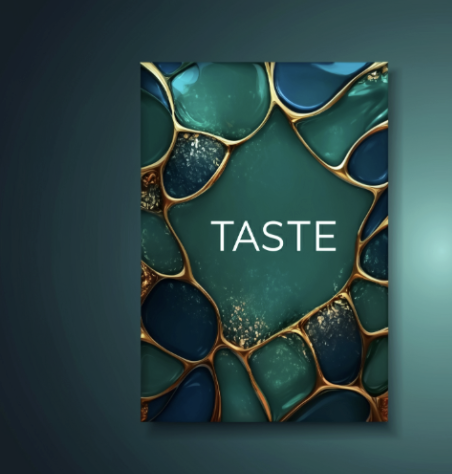}
    \caption*{(a) Winning Image ($x_w$)}
  \end{minipage}\hfill
  \begin{minipage}{0.42\textwidth}
    \includegraphics[width=\linewidth]{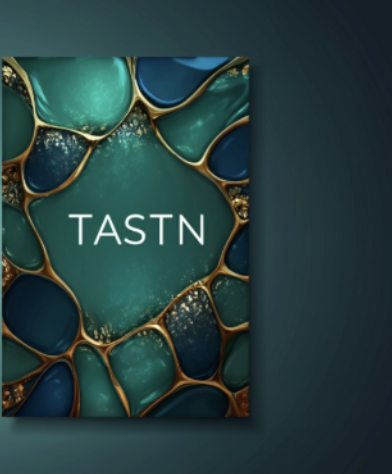}
    \caption*{(b) Losing Image ($x_l$)}
  \end{minipage}
  \caption{The split winning and losing images. The left image contains the correctly spelled word "TASTE", while the right contains the misspelling "TASTN".}
  \label{fig:passed_example}
\end{figure*}

\begin{strip}
\subsection{Example of a "Passed" Pair}
Figure \cref{fig:passed_example} shows an example of a high-quality pair that passed our filtering criteria. The backgrounds are identical, and the text differs by a single character, providing a clear and targeted signal for preference tuning.
\end{strip}

\begin{figure*}[ht]
  \centering
  \includegraphics[width=0.9\textwidth]{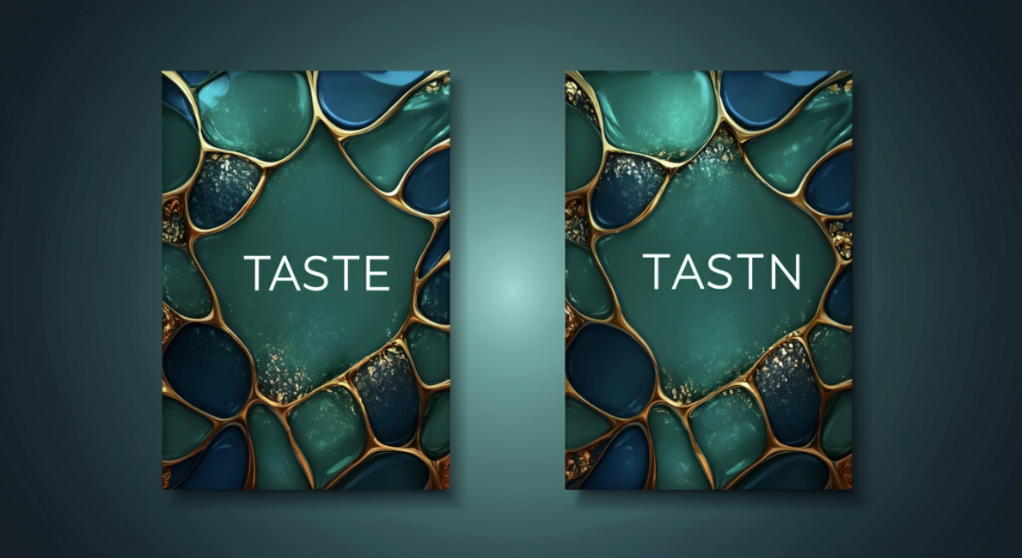}
  \caption{The original Diptych generated by Imagen.}
  \label{fig:passed_diptych}
\end{figure*}

\begin{figure*}[ht]
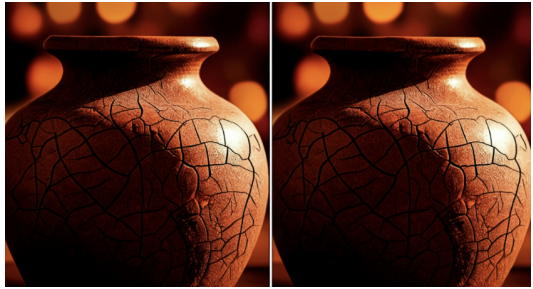

\begin{verbatim}
explanation: "Both images have the same background. The text is slightly
different; the first image says \"TASTE\" while the second image says
\"TASTN\". The difference in text is minor, and the rendering of the
text on the same background is consistent."
passing: true
confidence: 100
\end{verbatim}
\caption{Verifier response from a pair of passing images.}
\label{verification}
\end{figure*}

The verification model explanation for the passing pair is shown in \cref{verification}.

\begin{figure*}[h]
  \centering
  \includegraphics[width=0.4\textwidth]{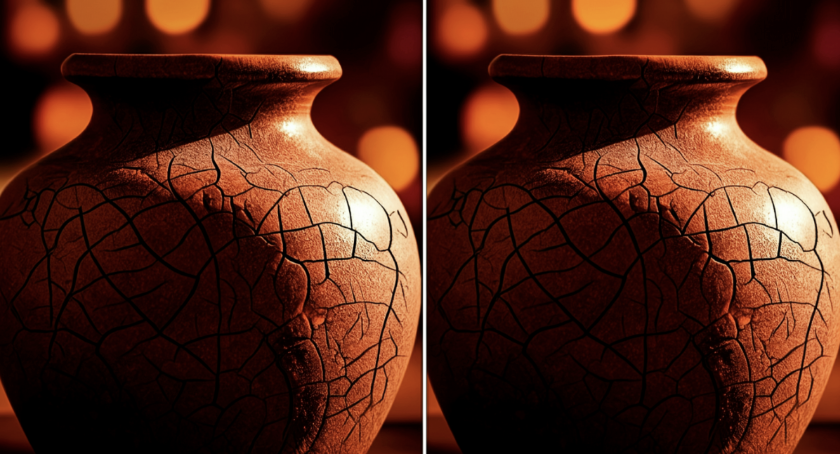}
  \caption*{
    \textbf{(a) Failure: No Text Rendered.} The model failed to render any text in either panel.
    \textit{Explanation: "Both images contain a picture of a cracked clay pot... There is no text present in either image."}
  }
  \vspace{1cm} 

  \includegraphics[width=0.4\textwidth]{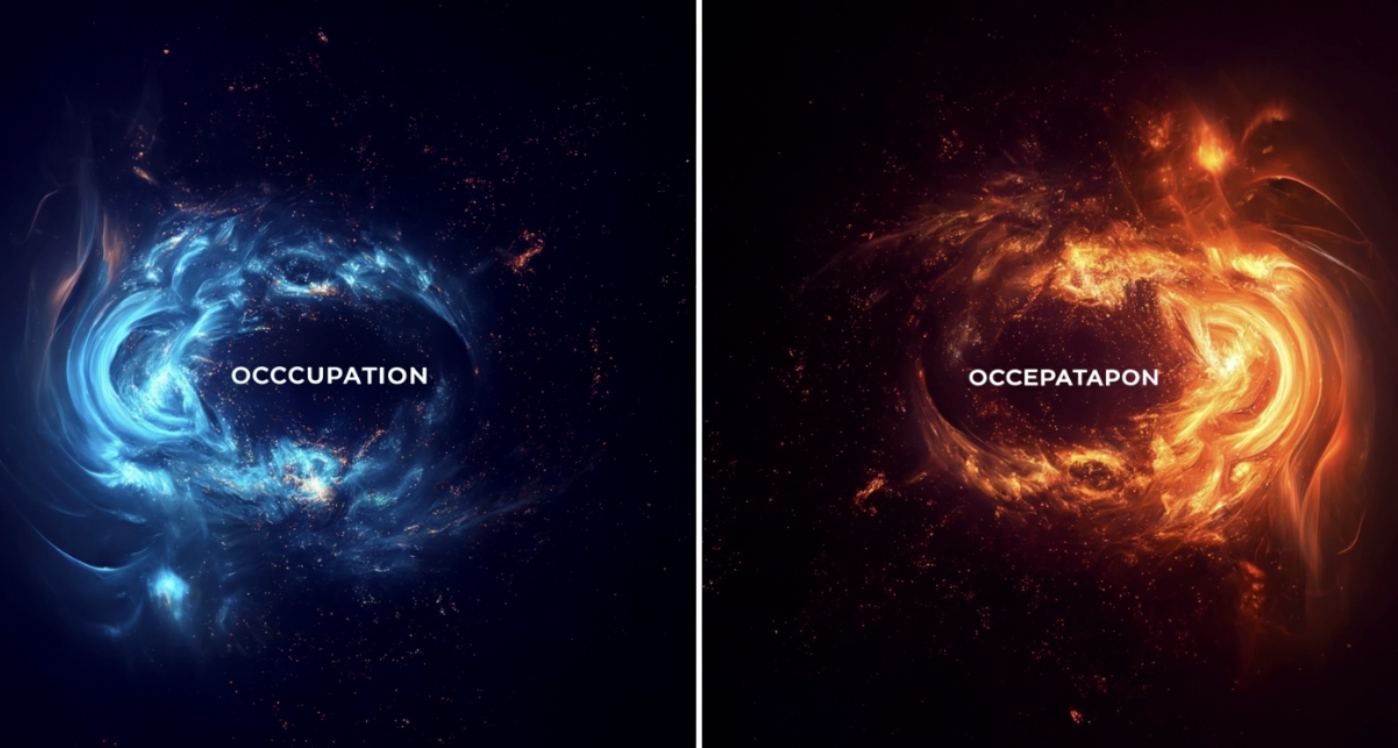}
  \caption*{
    \textbf{(b) Failure: Inconsistent Backgrounds.} Although the style is similar, the colors and details of the backgrounds differ significantly.
    \textit{Explanation: "The background, while sharing a similar style... has significantly different colors."}
  }
  \vspace{1cm}

  \includegraphics[width=0.4\textwidth]{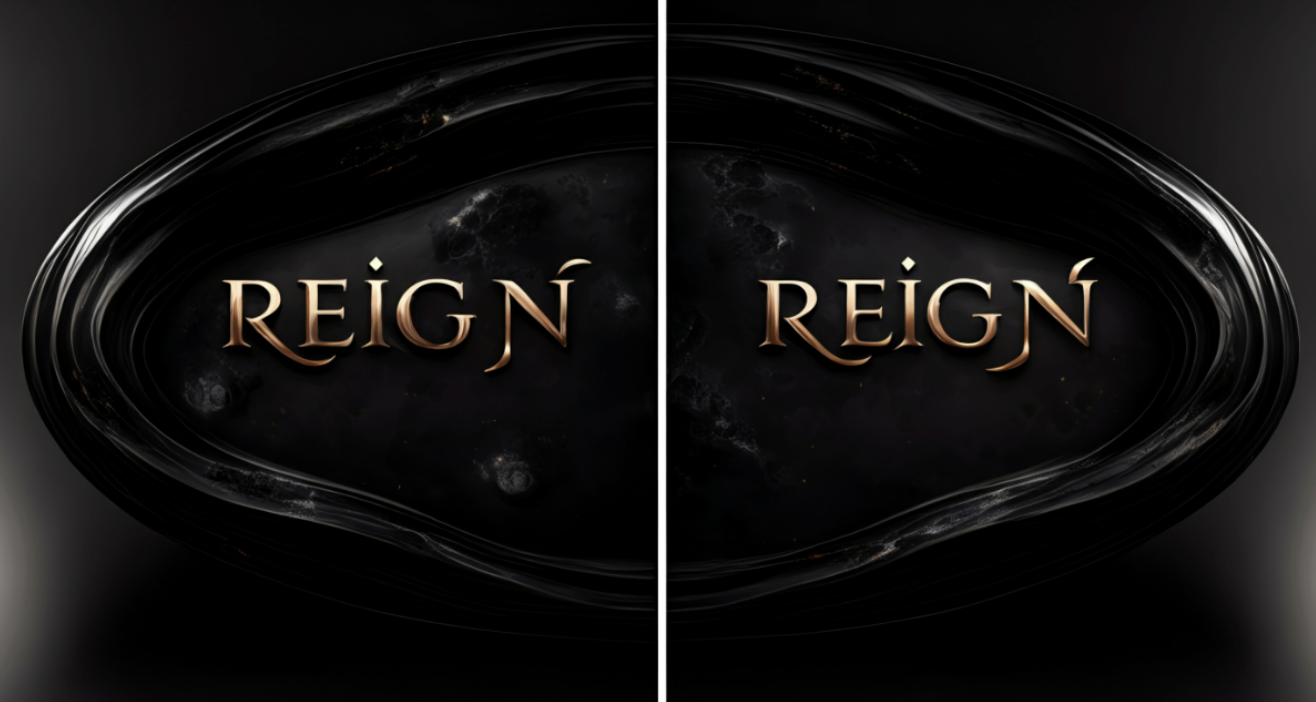}
  \caption*{
    \textbf{(c) Failure: Identical Text.} The model rendered the exact same text in both panels, failing to create a preference pair.
    \textit{Explanation: "The text 'REIGN' is present in both... The text itself is identical in both images..."}
  }
  
  \caption{Examples of generated pairs that were correctly identified as failures by our automated verification pipeline.}
  \label{fig:failed_examples}
\end{figure*}

\begin{strip}
\subsection{Examples of "Failed" Pairs}
Figure \cref{fig:failed_examples} illustrates common failure modes that our filtering process successfully catches. These examples are discarded and not used for training.
\end{strip}

\begin{strip}
\section{Sample Evaluation Prompts}
\label{sec:appendix_eval_data}

\begin{enumerate}
    \item A shipping label on a cardboard box that says 'Fragile - Handle With Care' and 'Contents: Glassware'
    \item A vibrant street art mural featuring the phrase 'Dream Big, Live Bold'
    \item A glossy magazine cover with the headline 'Future Forward' in large, striking typography
    \item A screenshot of a text message conversation, with one bubble saying 'Are you free tonight?' and the reply bubble saying 'Yes, I am! What's up?'
\end{enumerate}
\end{strip}

\end{document}